\documentclass[accepted]{uai2023} % for initial submission
\usepackage[american]{babel}
\usepackage{natbib} % has a nice set of citation styles and commands
\usepackage{mathtools} % amsmath with fixes and additions

\usepackage{hyperref}       % hyperlinks
\usepackage{url}            % simple URL typesetting
\usepackage{booktabs}       % professional-quality tables
\usepackage{nicefrac}       % compact symbols for 1/2, etc.
\usepackage{microtype}      % microtypography
\usepackage{xcolor}         % colors
\usepackage{graphicx}
\usepackage{subcaption}
\usepackage{float}
\usepackage{pgfplots}
\usepgfplotslibrary{fillbetween}
\usetikzlibrary{arrows}
\usetikzlibrary{backgrounds}
\usetikzlibrary{positioning}
\usepackage{amsmath}
\usepackage{amssymb}
\usepackage{mathtools}
\usepackage{bbm}
\usepackage{dsfont}
\newcommand{\bd}[1]{\boldsymbol{\mathrm{#1}}}
\newcommand{\R}{\mathbb{R}}
\newcommand{\LO}{\mathcal{O}}
\newcommand{\GP}{\mathcal{GP}}

\newcommand{\res}[2]{$#1$ \small $\pm #2$}
\newcommand{\xd}[1]{{\textcolor{blue}{{{}}#1}}{}}
\definecolor{highlightBlue}{RGB}{58,146,222}
\definecolor{highlightOrange}{RGB}{243,165,54}
\definecolor{highlightRed}{RGB}{208,2,27}
\definecolor{highlightGreen}{RGB}{65,117,5}
\definecolor{highlightPurple}{RGB}{144,19,254}
\definecolor{darkGray}{RGB}{50,50,50}
\definecolor{lightGray}{RGB}{200,200,200}

\title{Graph Classification Gaussian Processes via Spectral Features}

\author[1]{\href{mailto:<flo23@cam.ac.uk>?Subject=UAI2023: Graph Classification Gaussian Processes via Spectral Features}{Felix L. Opolka}{}}
\author[2]{Yin-Cong Zhi}
\author[1]{Pietro Li\`{o}}
\author[2]{Xiaowen Dong}
\affil[1]{%
    Department of Computer Science and Technology\\
    University of Cambridge\\
    Cambridge, UK
}
\affil[2]{%
    Department of Engineering Science\\
    University of Oxford\\
    Oxford, UK
}
  
\begin{document}
\maketitle

\begin{abstract}
Graph classification aims to categorise graphs based on their structure and node attributes.
In this work, we propose to tackle this task using tools from graph signal processing by deriving spectral features, which we then use to design two variants of Gaussian process models for graph classification.
The first variant uses spectral features based on the distribution of energy of a node feature signal over the spectrum of the graph.
We show that even such a simple approach, having no learned parameters, can yield competitive performance compared to strong neural network and graph kernel baselines.
A second, more sophisticated variant is designed to capture multi-scale and localised patterns in the graph by learning spectral graph wavelet filters, obtaining improved performance on synthetic and real-world data sets.
Finally, we show that both models produce well calibrated uncertainty estimates, enabling reliable decision making based on the model predictions.
\end{abstract}

\section{Introduction}\label{sec:introduction}

Data that are collected in a network environment, hence supported by a graph structure, have become pervasive in modern data analysis and processing tasks. This poses the new task of \textit{graph classification}, which, similar to image classification, aims at classifying graph-structured data into different classes. For example, representing protein structures as graphs, one may wish to classify whether they are toxic or not; modelling information propagation cascades on social media platforms as graphs, one may wish to detect whether the originating post of each cascade corresponds to fake news or not; considering urban transportation networks as graphs, one may wish to identify whether a particular area is likely to lead to traffic congestion.

Such graph-level classification problems are non-trivial generalisation of classical classification problems: graphs are irregular structures and traditional techniques defined in the Euclidean domains, such as the Fourier transform, cannot be applied directly; data collected on the nodes (or edges) of the graph are often continuous measurements, therefore traditional tools in graph analysis need to be adapted; finally, real-world graphs often come with extremely large size, making scalability of the algorithms a challenge.

Recent efforts to tackle the problem of graph classification mainly fall into two categories. First, graph kernels~\citep{nikolentzos2021graph, kriege2020survey, borgwardt2020graph}, as traditional ways of comparing graphs or compute distance between them, have been adapted for graph classification. However, it is still a challenge for these methods to handle multi-dimensional and continuous node features as well as graphs of different sizes. Second, graph neural networks~\citep{bruna2013spectral,defferrard2016chebnet,kipf2017semi}, generalisations of neural networks to deal with graph-structured data, can also be utilised for graph classification. In these frameworks, a read-out function is often deployed after the neural network layers to summarise node representations into a single graph representation, for example using summation, averaging, or pooling~\citep{dai2016discriminative, duvenaud2015convolutional, gilmer2017neural, ying2018hierarchical}. This addressed the issue of comparing graphs of different sizes; however, these architectures are often trained with large amount of data, and the predictions are not easily interpretable.

\begin{figure*}[t]
    \centering
    \begin{subfigure}[b]{0.24\linewidth}
    \includegraphics[width=\linewidth]{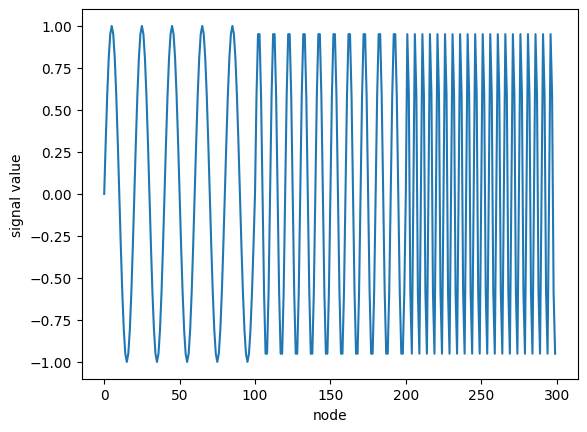}
    \caption{Graph signal $\bd{x}$ consisting of three concatenated sine-waves of different frequency.}\label{fig:graph_signal}
    \end{subfigure}\hfill
    \begin{subfigure}[b]{0.24\linewidth}
    \includegraphics[width=\linewidth]{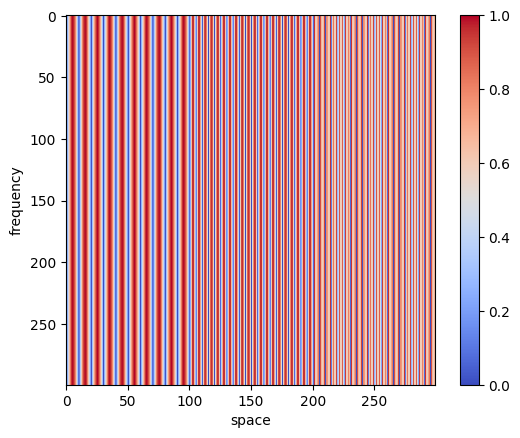}
    \caption{Signal under view \textbf{fully localised in space} (original signal $\bd{x}$)}\label{fig:spectogram_spatial}
    \end{subfigure}\hfill
    \begin{subfigure}[b]{0.24\linewidth}
    \includegraphics[width=\linewidth]{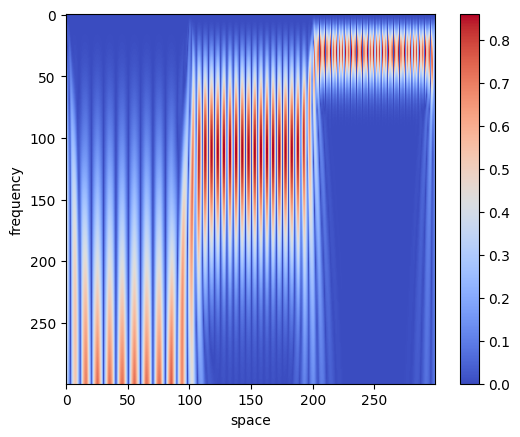}
    \caption{Signal under view \textbf{partially localised both in space and frequency} (Wavelet transform of $\bd{x}$)}\label{fig:spectogram_wavelet}
    \end{subfigure}\hfill
    \begin{subfigure}[b]{0.24\linewidth}
    \includegraphics[width=\linewidth]{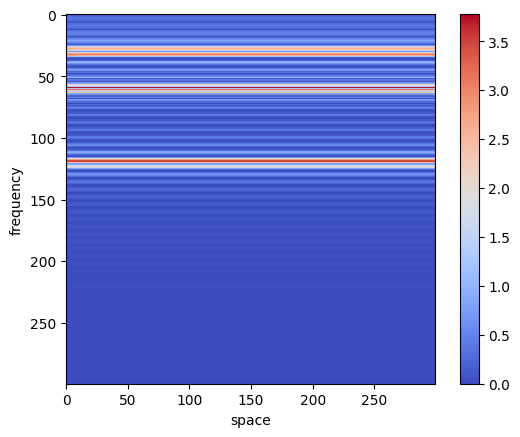}
    \caption{Signal under view \textbf{fully localised in frequency} (Fourier transform of $\bd{x}$)}\label{fig:spectogram_fourier}
    \end{subfigure}
    \caption{Visualisation of the different space-frequency resolutions of the spatial view of a signal (b), its wavelet transform (c), and its Fourier transform (d) for the example of concatenated sine-waves (a) on a path graph. A path graph forms a line of nodes where each node, except for the two end nodes, has exactly two neighbours.}\label{fig:spectograms}
\end{figure*}

We address the above limitations in this work, and our main contributions are as follows. First, inspired by the image segmentation literature~\citep{porter1996robust} as well as recent development in the field of graph signal processing~\citep{shuman2012emerging,ortega2018graph,dong2020graph,ortega2022introduction}, we propose to consider multi-dimensional and continuous node features as graph signals, and compute spectral features using the graph Fourier transform, i.e., energy distribution of the Fourier coefficients in different frequency bands. We then utilise these features in a Gaussian process framework, which has the advantage of not requiring a validation set and being more interpretable. We show that this simple method already achieves surprisingly competitive performance compared to baselines relying on graph kernels and graph neural networks. Second, we further derive a second method based on the spectral graph wavelets~\citep{hammond2011wavelets}, which possesses the additional benefits of capturing multi-scale and localisation information and leads to further improvement in classification performance. Finally, the proposed Gaussian process based methods allow us to quantify uncertainty information in the graph classification results, which to our knowledge has not been considered so far in the literature.

\section{Preliminaries}

The models proposed here will exploit spectral features of graphs within the framework of learning with Gaussian processes.
In the following, we will therefore give a brief introduction to the techniques from graph signal processing and Gaussian process inference that we will use in our methodology.

\subsection{Graph Signal Processing}\label{sec:gsp}

Graph signal processing (see \citet{ortega2022introduction} for a general introduction) offers a range of tools for analysing the spectral properties of graph signals.
One of its key tools is the generalisation of the Fourier transform to the graph domain, as described by~\citet{shuman2012emerging}.
% Our Gaussian process model for graph classification will draw heavily from graph signal processing, in particular the Fourier transform of graph signal \citep{shuman2012emerging} and spectral graph wavelets \citep{hammond2011wavelets} (see \citet{ortega2022introduction} for a general introduction).
% The Fourier transform is one of the most fundamental tools for analysing signals on the Euclidean domain.
It breaks down a signal into components of different frequency, giving rise to a complementary view of the signal on the frequency domain as opposed to the spatial domain.
Moreover, the Fourier transform is useful for analysing learned transforms of signals, which form the basis for machine learning.
Spectral graph theory \citep{chung1997spectral} has generalised the Fourier transform for signals on the Euclidean domain to signals on the more general graph domain.
In general, the Fourier transform of a signal on any domain is defined as the decomposition of that signal into the basis functions of the Laplace operator.
On the graph domain, the Laplace operator is given by the symmetric and positive-semidefinite Laplacian matrix $\bd{L} = \bd{D} - \bd{A}$, where $\bd{A} \in \R^{N \times N}$ is the adjacency matrix of the graph with $N$ nodes and $\bd{D} \in \R^{N \times N}$ is the diagonal degree matrix of the graph such that $D_{ii} = \sum_{j=1}^{N} A_{ij}$.
Using the eigendecomposition of the graph Laplacian $\bd{L} = \bd{U} \bd{\Lambda} \bd{U}^\top$, we can define the Fourier transform of a graph signal $\bd{x} \in \R^{N}$ as $\hat{\bd{x}} = \bd{U}^\top \bd{x}$.

The original signal $\bd{x}$ and its Fourier transform $\hat{\bd{x}}$ form the two extremes of a trade-off relationship between resolution in space and resolution in frequency: $\bd{x}$ is fully localised in space but not localised in frequency, whereas $\hat{\bd{x}}$ is fully localised in frequency but not localised in space.
The wavelet transform is a convenient technique for analysing signals with resolution in both space and frequency.
It decomposes a signal into a set of basis functions obtained by scaling and shifting a so-called \textit{mother wavelet}.
\citet{hammond2011wavelets} derive a graph signal wavelet transform for signal $\bd{x}$ using mother wavelet $b(\lambda)$ as
    \begin{equation}
    \bd{w} = \bd{U} b(\beta \bd{\Lambda}) \bd{U}^\top \bd{x},
    \end{equation}
where $\beta$ is the scale parameter. 
Adjusting this scale parameter allows examining the signal at varying frequency ranges, which correspond to different localisation behaviour in the spatial node domain.
% Equivalently, shifting the wavelet, i.e. localising it at different nodes, allows examining the signal at varying spatial ranges, i.e. locations in the graph.

The relationship between the original spatial signal, Fourier transform, and wavelet transform is best visualised for a signal on a path graph, i.e. a graph that simply forms a line of nodes, as it resembles an interval on the real line.
We plot a signal on such a graph consisting of three sine-waves of different frequencies in Figure~\ref{fig:graph_signal} along with the spectrograms of the signal in the spatial domain (\ref{fig:spectogram_spatial}), in the frequency domain (\ref{fig:spectogram_fourier}), and for the wavelet transform (\ref{fig:spectogram_wavelet}), which demonstrate how the wavelet transform trades off between spatial and spectral resolution.

To analyse a signal at multiple scales simultaneously, a low-pass filter $h(\alpha \lambda)$ with parameter $\alpha$ and multiple scaled versions of the mother wavelet can be combined into a more complex wavelet filter function $g_\theta(\lambda) = h(\alpha \lambda) + \sum_{l = 1}^{L} b(\beta_l \lambda)$ with the set of scale parameters $\theta = \{ \alpha, \beta_1, \beta_2, \ldots \}$.

\subsection{Gaussian Processes}\label{sec:gps}

Gaussian processes (see \citet{rasmussen2005gp} for a general introduction) are stochastic processes that can be considered multivariate normal distributions extended to infinitely many dimensions.
Their properties make them a convenient choice for prior distributions in Bayesian machine learning models.
A GP prior on a latent function $f: \R^{D} \rightarrow \R$ is given by
    \begin{equation}
    f(\bd{x}) \sim \GP\left( m(\bd{x}), k_\theta(\bd{x}, \bd{x}') \right),
    \end{equation}
where $m(\bd{x})$ is the mean function of the process and often set to 0 and $k_\theta(\bd{x}, \bd{x}')$ is its kernel function with a set of kernel hyperparameters $\theta$.
Given input data $\bd{X} = [\bd{x}_1, \bd{x}_2, \ldots, \bd{x}_N]^\top \in \R^{N \times D}$ and labels $\bd{y} = [y_1, y_2, \ldots, y_N]^\top \in \R$, the GP model can be used for performing probabilistic inference.
If the model specifies a Gaussian likelihood, the posterior distribution $p(\bd{f} \vert \bd{y})$ is analytically tractable and also follows a Gaussian process distribution.
Furthermore, the marginal likelihood $p(\bd{y})$ is tractable and can be maximised to optimise the kernel hyperparameters $\theta$.

Inference in a GP model becomes prohibitive for data sets of large size $N$, as computing the posterior distribution requires inverting an $N \times N$ covariance matrix.
\textit{Sparse Gaussian processes} alleviate this computational burden by constructing a smaller pseudo-data set of $M$ so-called \textit{inducing points}, where $M \ll N$.
The inducing points are chosen such that the GP posterior they induce is similar to the actual posterior using all $N$ data points.
\citet{titsias2009variational} proposes a way of learning the inducing inputs as variational parameters by optimising a lower bound to the marginal likelihood.

In case the model specifies a non-Gaussian likelihood such as a Categorical likelihood---as would be the case for classification tasks---the posterior distribution is no longer analytically tractable and needs to be approximated.
This can be achieved using a variational approximation to the posterior distribution where the variational family is chosen to be a Gaussian process.
\citet{hensman2015scalable} show how a variational approximation for non-Gaussian likelihoods can be designed for sparse GPs.
The resulting model has the benefit that the lower bound to the marginal likelihood, referred to as the \textit{Evidence Lower Bound (ELBO)}, has a data term that factorises over data points and can therefore be maximised using stochastic gradient descent~\citep{hensman2013gaussian}.

% * Similar to other kernel methods like Support Vector Machines (SVM), which are a popular choice for graph classification (see for example \citet{shervashidze2009efficient, shervashidze2011weisfeiler}), prior assumptions about the data are primarily embedded into the kernel function
% * A distinction of GPs, which influences kernel design, is that GPs allow optimising kernel hyperparameters
% * Moreover, GPs, unlike most other kernel methods, provide principled uncertainty estimates

\section{Spectral Feature Learning for Graph Classification}\label{sec:methodology}

% [Figure of time-frequency tiling of Fourier Transform and Wavelets; show that using optimised Wavelets, we can get better ,,view'' of the graph]
% * Plot spectrogram of learnt wavelets
%     * Same dataset, but different labels and compare spectograms afterwards

In the following, we present two GP models for graph classification.
The first model is focused on simplicity, while the second model trades off simplicity against higher expressive power.
The former is based on the graph Fourier transform of the node feature signal and is therefore referred to as \textbf{FT-GP}, whereas the latter employs the spectral graph wavelet transform and is therefore referred to as \textbf{WT-GP}.
The key idea in common for both approaches is that alternatives to the spatial view of the node feature graph signals, as provided by the Fourier and wavelet transform, are better starting points for designing graph kernels.
Consequently, we hypothesise that even simple transformations and representations of the signals under these views, mainly focusing on how the energy of these signals are distributed in different parts of the spectrum and/or space, are sufficiently expressive for distinguishing attributed graphs.

% We present a Gaussian process model for the graph-level prediction task based on spectral properties of graph signals.
% Two versions of the Gaussian process model are presented, a simpler baseline using the Fourier decomposition of the node feature signals, and a more advanced approach using spectral graph wavelets [maybe citation].

Both models are used for typical graph classification tasks where we assume to be given a set of $G$ training set graphs $\mathcal{T} = \{\mathcal{G}_1, \mathcal{G}_2, \ldots, \mathcal{G}_G\}$ of varying sizes $N_1, N_2, \ldots, N_G$, alongside their corresponding class labels $y_1, y_2, \ldots y_G$.
% Some of this introductory stuff might have to be moved to preliminaries
% * Each graph is given by a tuple $\mathcal{G}_i = (\mathcal{V}_i, \mathcal{E}_i)$
For each graph $i$, we are given an adjacency matrix  $\bd{A}^{(i)} \in \R^{N_i \times N_i}$ and $D$-dimensional node features $\bd{X}^{(i)} \in \R^{N_i \times D}$.
For the following exposition, we initially assume one-dimensional node features, i.e. $D=1$, and describe how to address the case of general $D$ later on.
The goal of graph classification is to learn a mapping from the adjacency matrix and node features of a graph to its class label such that it generalises to graphs outside the training set.

\subsection{Spectral features for graph-level prediction}

\begin{figure}[t]
    \subcaptionbox{Signal plotted for graph A\label{fig:enzyme_graph0}}[0.49\linewidth]{\includegraphics[width=\linewidth]{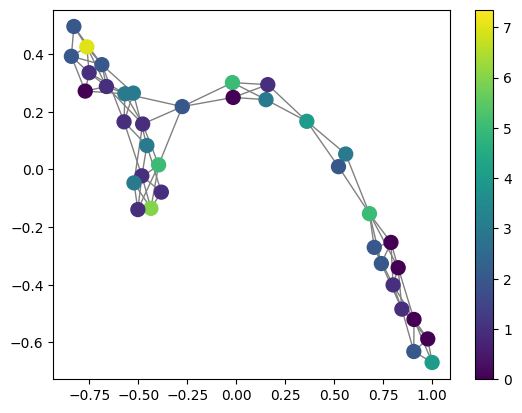}}
    \subcaptionbox{Cumulative energy function for graph A\label{fig:enzyme_cumulative0}}[0.49\linewidth]{\includegraphics[width=\linewidth]{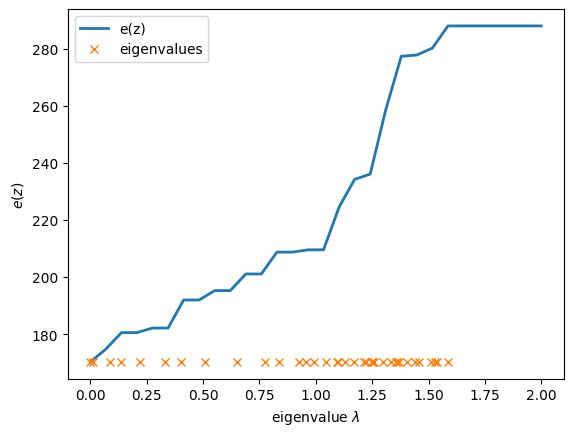}}
    
    \subcaptionbox{Signal plotted for graph B\label{fig:enzyme_graph1}}[0.49\linewidth]{\includegraphics[width=\linewidth]{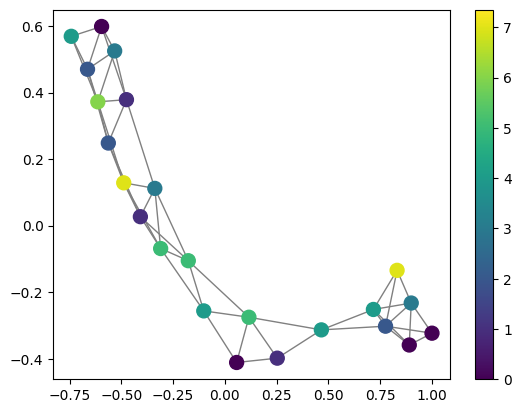}}
    \subcaptionbox{Cumulative energy function for graph B\label{fig:enzyme_cumulative1}}[0.49\linewidth]{\includegraphics[width=\linewidth]{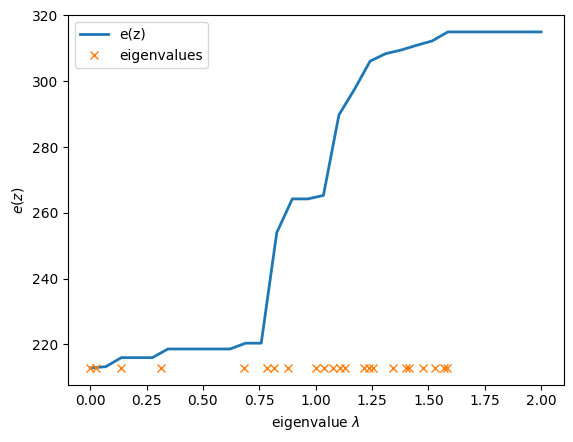}}
    \caption{Comparison of two graphs from the ENZYME data set~\citet{borgwardt2005protein} in terms of cumulative energy function. The same node feature is plotted for both graphs in (a) and (c) where the node colour indicates the node feature value. Figures (b) and (d) show the corresponding cumulative energy functions, revealing the distinct energy profiles of both graphs: in particular, graph A has a larger low-frequency component compared to graph B.}\label{fig:cumulative_energy}
\end{figure}

% The key motivation for our first approach is the insight that spectral signal processing can overcome challenges with learning from graphs of different sizes and simultaneously give rise to expressive graph kernels.
For each graph $\mathcal{G}^{(i)}$ with adjacency matrix $\bd{A}^{(i)} \in \R^{N_i \times N_i}$ %and one-dimensional node features $\bd{x}^{(i)} \in \R^{N_i}$ 
we can compute the symmetrically normalised graph Laplacian matrix $\bd{L}^{(i)} = \bd{I} - (\bd{D}^{(i)})^{-1/2}\bd{A}^{(i)}(\bd{D}^{(i)})^{-1/2}$.
Its eigendecomposition is $\bd{L}^{(i)} = \bd{U}^{(i)}\bd{\Lambda}^{(i)}\bd{U}^{(i)\top}$, where $\bd{U}^{(i)} \in \R^{N_i \times N_i}$ denotes the eigenvector matrix and $\bd{\Lambda}^{(i)} = \text{diag}(\lambda_1^{(i)}, \ldots, \lambda_{N_{i}}^{(i)}) \in \R^{N_i \times N_i}$ denotes the diagonal eigenvalue matrix.
The spectrum of the symmetrically normalised graph Laplacian is in $[0, 2]$, i.e. $0 \leq \lambda_j^{(i)} \leq 2 \,\forall j = 1, \ldots, N_i$ \citep{shuman2012emerging}.
Notably, this holds regardless of the size $N_i$ of the graph.

Using the definition of the graph Fourier transform as presented in Section \ref{sec:gsp}, we can compute the Fourier coefficients of the node feature signal for graph $\mathcal{G}^{(i)}$ as $\hat{\bd{x}}^{(i)} = \bd{U}^{(i)\top} \bd{x}^{(i)}$.
Based on the eigenvalues of the graph and the Fourier coefficients of the node feature signal, we can define a \textit{cumulative energy function} for graph $i$ as
    \begin{equation}
    e^{(i)}(z) = \sum\limits_{j = 1}^{N^{(i)}} \hat{x}^{(i)2}_{j} \mathds{1}_{\{\lambda_j^{(i)} \leq z\}}.
    \end{equation}
This function $e^{(i)}(z)$ represents the energy of the node feature signal that is contained in the spectrum up to a frequency $z$ and it thereby encodes both graph structure and node feature information. 
We hypothesise that the energy profile formulated in this way is expressive enough to distinguish attributed graphs of different classes.
Figure~\ref{fig:cumulative_energy} visualises the cumulative energy function for two sample graphs from the ENZYMES data set~\citet{borgwardt2005protein}.
We find that the two cumulative energy functions differ clearly, both as a result of different eigenvalue locations and distinct Fourier coefficients.

We can derive a feature vector $\bd{e}^{(i)} \in \R^{M}$ for graph $i$ from $e^{(i)}(z)$ by evaluating $e^{(i)}(z)$ at a sequence of $M$ evaluation points $[h_1, \ldots, h_M]$, therefore
    \begin{equation}
    e^{(i)}_m = e^{(i)}(h_m) = \sum\limits_{j = 1}^{N^{(i)}} \hat{x}^{(i)2}_{j} \mathds{1}_{\{\lambda_j^{(i)} \leq h_m\}}.
    \end{equation}
In practice, we will often choose $[h_1, \ldots, h_M]$ to be linearly spaced on the $[0, 2]$ interval of the spectrum.
Crucially, this guarantees that all graph representations $\bd{e}^{(1)}, \ldots, \bd{e}^{(G)}$ are of size $M$ regardless of individual graph sizes.

The above derivation assumes one-dimensional node features ($D = 1$).
The graph representations can be generalised to multi-dimensional node features ($D > 1$) by defining a cumulative energy function for each feature dimension and concatenating their discretisations, resulting in graph representations $\bd{e}^{(i)}$ of size $M \times D$.

The resulting graph representations can now serve as input into a base kernel of choice and we use the radial basis function (RBF) kernel $k(\bd{x}, \bd{x}') = \exp(- \frac{l}{2} \lVert\bd{x} - \bd{x}'\rVert_2)$ with lengthscale $l$.

\subsection{Spectral wavelet features for graph-level prediction}

A key limitation of the FT-GP approach stems from the Fourier transform providing full resolution in frequency but no resolution in space.
As a result, we expect the model to under-perform in correctly classifying graphs whose class is determined by localised patterns (cf. Figure~\ref{fig:spectograms}).
This is evaluated in more detail in experiments with synthetic data in Section~\ref{sec:synthetic}.
We can partially alleviate this limitation by employing the wavelet transform to derive kernels, allowing us to better trade off between localisation in space and frequency in a flexible manner. 
% * limitation of previous approach is that basis functions of Fourier decomposition are fully localised in frequency domain but not localised in spatial domain
% * will therefore be unable to correctly classify graphs whose class is determined by localised patterns (cf. Figure~\ref{fig:spectograms})
%     * [refer to synthetic experiments]
% * this can be partially alleviated by employing Wavelet basis functions to derive kernels allowing to better trade off between localisation in space and frequency in a flexible manner

\begin{figure}[t]
    \subcaptionbox{Wavelet filter A\label{fig:wavelet_filter1}}[0.46\linewidth]{\includegraphics[width=\linewidth]{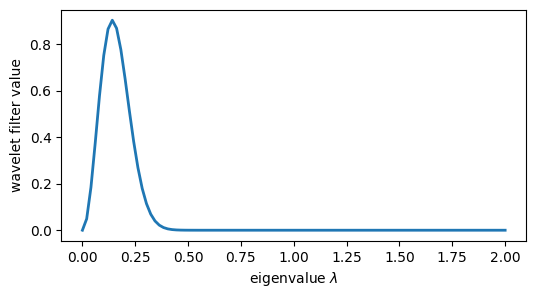}}\hspace{0.38cm}
    \subcaptionbox{Wavelet filter B\label{fig:wavelet_filter2}}[0.46\linewidth]{\includegraphics[width=\linewidth]{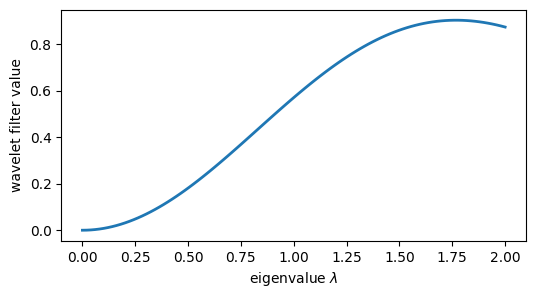}}
    
    \subcaptionbox{Wavelet atom localised at red node (filter A)\label{fig:wavelet_coeffs1}}[0.49\linewidth]{\includegraphics[width=\linewidth]{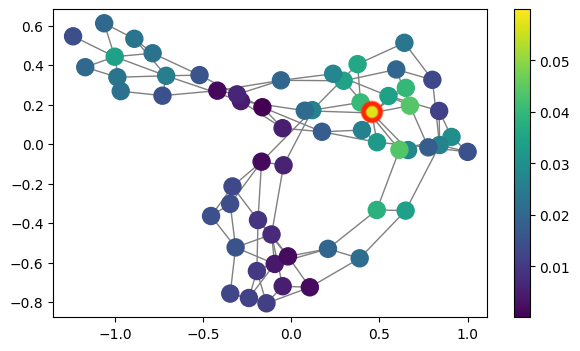}}
    \subcaptionbox{Wavelet atom localised at red node (filter B)\label{fig:wavelet_coeffs2}}[0.49\linewidth]{\includegraphics[width=\linewidth]{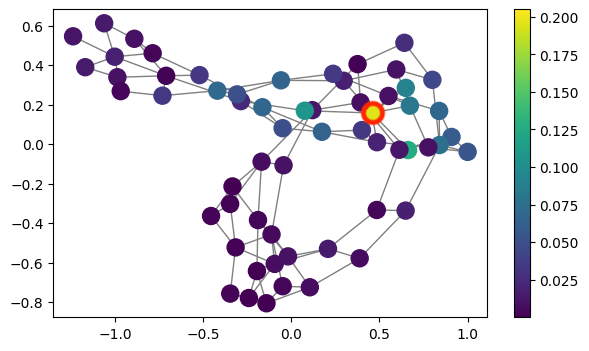}}
    
    \subcaptionbox{Signal filtered with filter A\label{fig:wavelet_filtered1}}[0.49\linewidth]{\includegraphics[width=\linewidth]{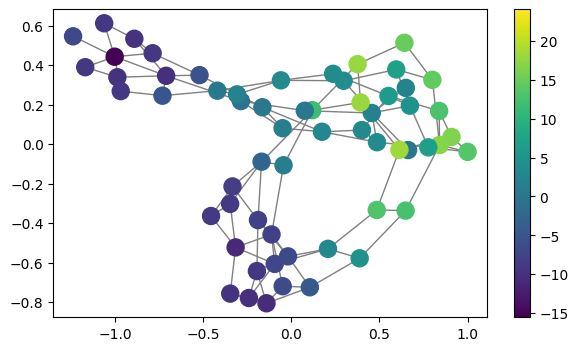}}
    \subcaptionbox{Signal filtered with filter B\label{fig:wavelet_filtered2}}[0.49\linewidth]{\includegraphics[width=\linewidth]{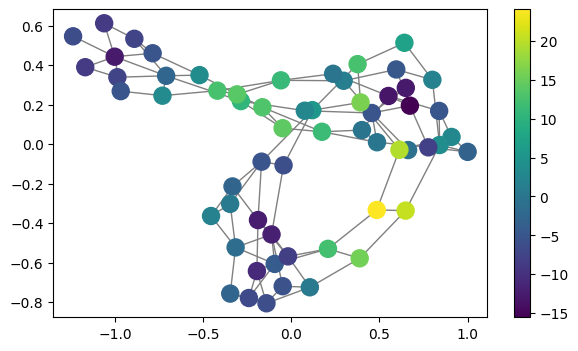}}
    \caption{Comparison of two different filters for the same graph and node feature (as a graph signal) from the ENZYME data set. The band-pass filter of large wavelet scale (a) highlights low frequencies, captures a larger node neighbourhood (c), and produces a smooth signal (e). The band-pass filter of small wavelet scale (b) highlights high frequencies, captures a smaller node neighbourhood (d), and produces a less smooth signal (f). %Both offer a distinct view of the signal on the graph.
    }\label{fig:wavelets}
\end{figure}

We design wavelet filters consisting of a single low-pass filter and multiple band-pass filters to obtain filtered signals.
A particular wavelet filter offers a distinct view of the attributed graph at hand, hence using $K$ of those wavelet filters allows us to obtain a more diverse view of the graph.
Moreover, we can seize the capability of GPs to optimise hyper-parameters to find wavelet scales that better distinguish the classes of graphs of the particular data set at hand.
% * Use a wavelet filter consisting of a low-pass filter and multiple band-bass filters to compute wavelet transform of node feature signal (cf. Figure~\ref{fig:spectogram_wavelet})
% * Each specific Wavelet filter offers a particular view of the attributed graph at hand
% * Use multiple of those wavelet filters to obtain a more diverse view of the attributed graph at hand
% * We can use the GPs capability to optimise hyper-parameters to optimise the scales of the wavelet filters to better distinguish the classes of graphs of the particular data set at hand
Figure~\ref{fig:wavelets} plots the wavelet transformed signal for a particular graph and two filters of different scales, to showcase the distinct view of the attributed graph each filter offers.

We define each of the $K$ wavelet filter functions as a sum of atomic filters including a single low-pass filter and $L$ band-pass filters
    \begin{equation}
    g_\theta(\lambda) = h(\alpha \lambda) + \sum\limits_{l = 1}^{L} b(\beta_l \lambda),
    \end{equation}
where $h(\alpha\lambda)$ refers to the low-pass filter function with scale $\alpha$, $b(\beta_l\lambda)$ refers to the $l$-th band-pass filter with scale $\beta_l$, and $\theta = \{ \alpha, \beta_1, \beta_2, \ldots \}$ is the set of scale parameters.
Reminiscent of the signal energy computed for the FT-GP model proposed earlier, we compute the magnitude of signal filtered with filter $k$
    \begin{equation}
    w^{(i)}_k = \lVert \bd{U}^{(i)} g_{\theta_k}(\bd{\Lambda}^{(i)}) \bd{U}^{(i)\top} \bd{x}^{(i)} \rVert_2
    \end{equation}
to express how much energy of the signal is captured by the $k$th filter.
The final feature vector $\bd{w}^{(i)} \in \R^{K \cdot D}$ for graph $i$ stacks the filter coefficients for each of the $K$ wavelet filters and each of the $D$ node feature signals.
As before, the resulting signal representations are fed into a radial basis function kernel to obtain the covariance between graphs.
The kernel hyper-parameters are optimised alongside the wavelet scale parameters $\theta$ in a data-driven way employing type II maximum likelihood estimation.

\subsection{Relationship to graph neural networks}\label{sec:relationship}

The GP models presented here operate in two stages, one focused on transformations of the node feature signal and the second on deriving a graph-level summary.
In this regard, they share similarities with graph neural networks for graph classification, making it worthwhile to compare the two approaches more thoroughly.

Both GP methods, in the first step, translate the node feature signal from the spatial domain to a domain that is either fully localised in frequency or partially localised in both space and frequency.
The resulting coefficients thereby combine information from all nodes across the graph, overcoming the spatial locality of each node.
In case of WT-GP, the wavelet transform can be considered an aggregation operation of node features in a neighbourhood around each node whose shape is determined by the particular wavelet filter in use~\citep{opolka2022adaptive}.
This aggregation of information across graphs is reminiscent of the aggregation operation inherent to graph convolutional networks (GCN)~\citep{kipf2017semi}.
In fact, \citet{wu2019simplifying} describe how the GCN aggregation operation acts as a low-pass filter on the node feature graph signal.

In a second step, the GP methods aggregate coefficients across the whole graph via binning in case of FT-GP and summation in case of WT-GP.
Neural network methods apply similar so-called ``read out'' operations to summarise node representations into a single graph representation via simple summation or averaging~\citep{dai2016discriminative, duvenaud2015convolutional, gilmer2017neural}, as well as more complicated pooling operations~\citep{ying2018hierarchical}.
In contrast to GCN-based approaches however, the aggregation approach presented here has a straightforward spectral interpretation and allows aggregating information in a way that goes beyond averaging low-pass filtered values, making it possible to capture multi-scale information in the graphs.
Furthermore, the GP framework provides the flexibility of a non-parametric approach and outputs uncertainty estimates for predictions, which are examined in more detail in Section~\ref{sec:uncertainty}.

\subsection{Scalability}

In terms of scalability of the GP models, the quantities of concern are the number of graphs in a data set and the size of those graphs.
Within the computation of the kernel of either proposed approach, the computational complexity is dominated by the necessary eigendecomposition of the graph Laplacian, which is in $\LO(N_i^3)$, where $N_i$ is the size of a graph $i$.
Graphs in graph classification tasks are typically small (cf. Table~\ref{tab:data_stats}), therefore the number of nodes is not usually an obstacle for scalability.
For larger graphs, one can reduce the computational complexity by resorting to the approximate Fourier transform of a graph signal~\citep{lemagoarou2018approximate}.
% Furthermore, both kernel formulations allow pre-computing the eigenvalues, eigenvectors, and Fourier transformed signals, which reduces the computational cost during learning.

The total number of graphs in the data set may pose another limit to the scalability of the method when computing the GP posterior.
Sparse GPs as discussed in Section~\ref{sec:gps} alleviate this issue by enabling stochastic optimisation in mini-batches when the data set size requires it.

\section{Related Work}

\textbf{Applications of Spectral and Wavelet Energy:} Our work is related to building classifiers in the graph spectral domain, and wavelets are an extension to such approaches with the benefits of multi-scale properties and better localisation. Solving problems on graphs by making use of the energy distribution in the spectral domain has been demonstrated effectively for the task of image segmentation in classical signal processing~\citep{porter1996robust}, and for geographical mobility prediction in graph signal processing~\citep{dong2013inference}. Both suggest that the energy distribution clearly contains a significant level of information which we aim to utilise in this work for graph classification.

\textbf{Graph Kernels:} One of the first studies into kernel functions acting on the graph level is the graph kernels summarised in the surveys of \citet{nikolentzos2021graph, kriege2020survey, borgwardt2020graph}. Graph kernels generally have multiple definitions, of which a number of choices have in common the existence of a double sum of a base kernel. To compute the kernel between two graphs, a standard base kernel is chosen and takes as inputs a node feature from each graph, and the double summation then covers all possible pairs of nodes between the two graphs. Similar definition can be applied to graphs that contain edge features. Such designs are limited as any edge connections are ignored in the double sum, and the kernel therefore boils down to a comparison of cross-products of the node features. Another common design is the computation of a product graph, where a new graph is constructed based on the two input graphs. When we simulate a random walk on the product graph, we compute the number of matching walks on the two individual graphs. Random walk kernels do take into account the graph structure, but is limited by scalability as the computation of the kernel over a product graph leads to a $\mathcal{O}(N_i^3N_j^3)$ complexity.

Graph kernels has also been developed from the Weisfeiler-Lehman (WL) test for graph isomorphism \citep{huang2021short}, which also acts as a graph similarity. The WL test produces a series of node ``colouring'' from a neighbourhood gathering and relabelling procedure. In the WL kernel \citep{shervashidze2011weisfeiler}, the ``colourings'' are passed through a hashing function (or histogram mapping) to form inputs to a base kernel. As an alternative, the Wasserstein distance between two sets of labels can be used instead of the hashing function as shown in \citet{togninalli2019wasserstein}. Compared to our kernel design, WL-based kernels generally work with the neighbourhoods in the spatial domain, and so they ignore any information in the spectral domain. Additionally, if the graph contains high dimensional node features, the WL does not have an efficient way to encode them into the embeddings for the WL algorithm.

The use of Fourier and wavelets basis are Laplacian-based models. Wavelets in particular, allow for aggregation over a continuous neighbourhood, giving the model the ability to operate on a multi-scale level. Though multi-scale Laplacian kernels exist such as \citet{kondor2016multiscale}, they only operate in the spatial domain by computing kernels between sub-graphs of various neighbourhood sizes. On the other hand, the work of \citet{pineau2019using} does make use of the Laplacian spectral information, but like the previous kernel, it ignores the node features that may come with the graphs, while we focus on the spectral information of the node features.

\textbf{Graph Neural Network Models:} Lastly, graph neural networks (GNNs) have also been applied as a test for graph isomorphism, and as a result can be applied to graph classification. The message passing step of a GNN layer (examples including \citet{duvenaud2015convolutional, li2012nested, murphy2019relational}) is comparable to the neighbourhood gathering step in the WL algorithm. Analysis of certain GNN models showed that they are at best as powerful as the WL test, and the graph isomophism network (GIN) proposed in \citet{xu2019powerful} achieves the theoretical guarantee of the WL. Meanwhile, simpler models such as GCN \citep{kipf2017semi} and GraphSAGE \citep{hamilton2017inductive} have been shown to be unable to distinguish certain types of graphs that GIN can handle.

\iffalse
Potential work to mention:

* Multi-scale Laplacian Kernel \cite{kondor2016multiscale}

* WL kernel \cite{shervashidze2011weisfeiler} and the WL test \cite{huang2021short}

* GIN \cite{xu2019powerful} and maybe also GraphSAGE \cite{hamilton2017inductive}

* Graph kernels from surveys \cite{nikolentzos2021graph, kriege2020survey}

\xd{could also mention:\\
- classical signal processing papers on wavelet band energies\\
- graph signal processing papers on Fourier/wavelet energy distributions
}
\fi

\section{Experiments}

Our work aims to investigate two core empirical questions with regards to the models presented.
Firstly, whether the energy profile as captured by either of the presented kernels is sufficiently expressive to classify real-world graphs.
Secondly, whether the wavelet transform approach of WT-GP improves upon the FT-GP model based on the Fourier transform.
We begin with the latter question by comparing the two proposed models on synthetic data sets.

In all our experiments, we use the same experimental setup, unless explicitly stated otherwise.
The FT-GP discretises the cumulative energy function by evaluating it at $30$ evaluation points.
The WT-GP uses $10$ filters consisting of a single low-pass and three band-pass filters. 
Each low-pass scale of each filter is uniformly randomly initialised to a value between $4.0$ and $6.0$.
The band-pass scales of each filter are uniformly randomly initialised to a value between $0.1$ and $5.0$.
A non-sparse variational Gaussian process as described in Section~\ref{sec:gps} is trained using the L-BFGS-B optimiser~\citep{liu1989limited, byrd1995limited} until the ELBO convergences.
All results are cross-validated using 10-folds under a stratified split where in each fold $80\%$ of the data is used for training and $10\%$ for testing.
The remaining $10\%$ are set aside for validation to make the evaluation comparable to results in related work, although neither of the two GP models make use of the validation set.
This validation procedure is suggested by~\citet{errica2020fair} to overcome weaknesses in the evaluation of graph classification methods in previous work.

\subsection{Synthetic experiments}\label{sec:synthetic}

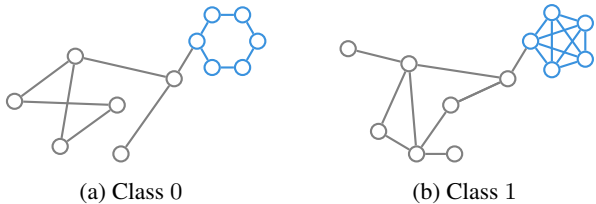
\begin{figure}[t]
    \centering
    \begin{subfigure}[b]{0.4\linewidth}
        \centering
        \begin{tikzpicture}
        \node[circle, draw=gray, thick, inner sep=0, minimum width=0.2cm] (v0_1) at (0.0, 0.0) {};
        \node[circle, draw=gray, thick, inner sep=0, minimum width=0.2cm] (v0_2) at (1.3, -0.3) {};
        \node[circle, draw=gray, thick, inner sep=0, minimum width=0.2cm] (v0_3) at (-0.8, -0.6) {};
        \node[circle, draw=gray, thick, inner sep=0, minimum width=0.2cm] (v0_4) at (0.55, -0.65) {};
        \node[circle, draw=gray, thick, inner sep=0, minimum width=0.2cm] (v0_5) at (-0.2, -1.2) {};
        \node[circle, draw=gray, thick, inner sep=0, minimum width=0.2cm] (v0_6) at (0.6, -1.3) {};
        \draw[gray, thick] (v0_1) -- (v0_2);
        \draw[gray, thick] (v0_1) -- (v0_3);
        \draw[gray, thick] (v0_1) -- (v0_5);
        \draw[gray, thick] (v0_2) -- (v0_6);
        \draw[gray, thick] (v0_3) -- (v0_4);
        \draw[gray, thick] (v0_4) -- (v0_5);
    
        \node[circle, draw=highlightBlue, thick, inner sep=0, minimum width=0.2cm] (r0_1) at (1.6, 0.2) {};
        \node[circle, draw=highlightBlue, thick, inner sep=0, minimum width=0.2cm] (r0_2) at (1.8, 0.55) {};
        \node[circle, draw=highlightBlue, thick, inner sep=0, minimum width=0.2cm] (r0_3) at (2.2, 0.55) {};
        \node[circle, draw=highlightBlue, thick, inner sep=0, minimum width=0.2cm] (r0_4) at (2.4, 0.2) {};
        \node[circle, draw=highlightBlue, thick, inner sep=0, minimum width=0.2cm] (r0_5) at (2.2, -0.15) {};
        \node[circle, draw=highlightBlue, thick, inner sep=0, minimum width=0.2cm] (r0_6) at (1.8, -0.15) {};
        \draw[gray, thick] (v0_2) -- (r0_1);
        \draw[highlightBlue, thick] (r0_1) -- (r0_2);
        \draw[highlightBlue, thick] (r0_2) -- (r0_3);
        \draw[highlightBlue, thick] (r0_3) -- (r0_4);
        \draw[highlightBlue, thick] (r0_4) -- (r0_5);
        \draw[highlightBlue, thick] (r0_5) -- (r0_6);
        \draw[highlightBlue, thick] (r0_6) -- (r0_1);
        \end{tikzpicture}
        \caption{Class $0$}
    \end{subfigure}\hspace{1cm}
    \begin{subfigure}[b]{0.4\linewidth}
        \centering
        \begin{tikzpicture}
        \node[circle, draw=gray, thick, inner sep=0, minimum width=0.2cm] (v0_1) at (0.0, -0.1) {};
        \node[circle, draw=gray, thick, inner sep=0, minimum width=0.2cm] (v0_2) at (1.3, -0.3) {};
        \node[circle, draw=gray, thick, inner sep=0, minimum width=0.2cm] (v0_3) at (-0.8, 0.1) {};
        \node[circle, draw=gray, thick, inner sep=0, minimum width=0.2cm] (v0_4) at (0.55, -0.65) {};
        \node[circle, draw=gray, thick, inner sep=0, minimum width=0.2cm] (v0_5) at (-0.4, -1.0) {};
        \node[circle, draw=gray, thick, inner sep=0, minimum width=0.2cm] (v0_6) at (0.6, -1.3) {};
        \node[circle, draw=gray, thick, inner sep=0, minimum width=0.2cm] (v0_7) at (0.1, -1.3) {};
        \draw[gray, thick] (v0_1) -- (v0_2);
        \draw[gray, thick] (v0_1) -- (v0_3);
        \draw[gray, thick] (v0_1) -- (v0_7);
        \draw[gray, thick] (v0_1) -- (v0_5);
        \draw[gray, thick] (v0_2) -- (v0_4);
        \draw[gray, thick] (v0_2) -- (v0_4);
        \draw[gray, thick] (v0_4) -- (v0_7);
        \draw[gray, thick] (v0_5) -- (v0_7);
        \draw[gray, thick] (v0_6) -- (v0_7);

        % x position: cos(180 degrees -72*4 degrees) * 0.4 + 2.0
        % y position: sin(180 degrees -72*4 degrees) * 0.4 + 0.2
        \node[circle, draw=highlightBlue, thick, inner sep=0, minimum width=0.2cm] (r0_1) at (1.6, 0.2) {};
        \node[circle, draw=highlightBlue, thick, inner sep=0, minimum width=0.2cm] (r0_2) at (1.88, 0.58) {};
        \node[circle, draw=highlightBlue, thick, inner sep=0, minimum width=0.2cm] (r0_3) at (2.32, 0.44) {};
        \node[circle, draw=highlightBlue, thick, inner sep=0, minimum width=0.2cm] (r0_4) at (2.32, -0.04) {};
        \node[circle, draw=highlightBlue, thick, inner sep=0, minimum width=0.2cm] (r0_5) at (1.88, -0.18) {};
        \draw[gray, thick] (v0_2) -- (r0_1);
        \draw[highlightBlue, thick] (r0_1) -- (r0_2);
        \draw[highlightBlue, thick] (r0_1) -- (r0_3);
        \draw[highlightBlue, thick] (r0_1) -- (r0_4);
        \draw[highlightBlue, thick] (r0_1) -- (r0_5);
        \draw[highlightBlue, thick] (r0_2) -- (r0_3);
        \draw[highlightBlue, thick] (r0_2) -- (r0_4);
        \draw[highlightBlue, thick] (r0_2) -- (r0_5);
        \draw[highlightBlue, thick] (r0_3) -- (r0_4);
        \draw[highlightBlue, thick] (r0_3) -- (r0_5);
        \draw[highlightBlue, thick] (r0_4) -- (r0_5);
        \end{tikzpicture}
        \caption{Class $1$}
    \end{subfigure}
    \caption{Visualisation of example graphs of the two different classes in the synthetic ring-vs-clique data set. All graphs consist of a graph component sampled from an Erd\H{o}s-R\'{e}nyi (ER) model, which is connected to either a ring graph component in case of class $0$ or a fully connected graph component in case of class $1$ (highlighted in \textcolor{highlightBlue}{blue}). Both components of each graph can vary in the number of nodes.}
    \label{fig:ring-vs-clique}
\end{figure}

We design two synthetic data sets to highlight the differences between the two GP models presented here.
The first data set, referred to as \textbf{ring-vs-clique}, contains two classes. 
The classes differ in that graphs in class $0$ are guaranteed to contain a subgraph that forms a ring of at least size $5$, while graphs in class $1$ are guaranteed to contain a subgraph that forms a complete graph (or clique) of at least size $5$.
We generate the class-balanced data set by sampling $200$ graphs using the following procedure. 
In a first step, a graph of uniformly random size between $10$ and $30$ nodes is sampled from an Erd\H{o}s-R\'{e}nyi model~\citep{erdos1959random}.
In the second step, depending on the class label, a ring or clique of uniformly random size between $5$ and $10$ nodes is constructed and connected to a randomly chosen node in the graph sampled in the first step through a single edge. The two different classes are sketched in Figure~\ref{fig:ring-vs-clique}.
The data set is designed to test how well a model can distinguish graphs when class labels are determined only by a relatively small, localised subgraph.

The second data set, referred to as \textbf{sbm}, contains graphs drawn from a stochastic block model~\citep{holland1983stochastic}, where nodes of the same block are connected with a probability of $80\%$ and nodes of different blocks are connected with $10\%$ probability. Graphs of class $0$ are drawn from a model with $2$ blocks and graphs of class $1$ are drawn from a model with $3$ blocks. All graphs have a uniform random size between $10$ and $30$ nodes.

The results for the binary classification tasks on these data sets for the two GP models are shown in Table~\ref{tab:synthetic_results}.
We find that on both tasks, the wavelet-based GP model strongly outperforms the baseline FT-GP model.
In fact, FT-GP performs only slightly better than random guessing on both data sets whereas WT-GP achieves a near perfect accuracy on ring-vs-clique and a high accuracy on the sbm data set.
The results confirm that designing graph kernels based on wavelet filtered node feature signals leads to more expressive kernels compared to a kernel based on Fourier transformed node features.

\begin{table}
    \centering
    \begin{tabular}{lcc}
    \toprule
     & \textbf{ring-vs-clique} & \textbf{sbm} \\
     \midrule
    \textbf{FT-GP} & \res{62.5}{7.5} & \res{58.5}{15.2} \\
    \textbf{WT-GP} & \res{\bd{99.5}}{1.5} & \res{\bd{91.0}}{5.4} \\
    \bottomrule
    \end{tabular}
    \caption{Classification accuracy of the FT-GP model compared to the WT-GP model on the two synthetic binary classification data sets.}
    \label{tab:synthetic_results}
\end{table}

\begin{table}
    \centering
    \resizebox{\linewidth}{!}{
    \begin{tabular}{lccccc}
    \toprule
    \textbf{Data set} & \textbf{\# Graphs} & \textbf{\# Classes} & \textbf{Avg \# Nodes} & \textbf{Avg \# Edges} & \textbf{\# Node Attr} \\
    \midrule
    \textbf{ENZYMES} & $600$ & $6$ & $32.63$ & $62.14$ & $21$ \\
    \textbf{MUTAG} & $188$ & $2$ & $17.93$ & $19.79$ & $7$ \\
    \textbf{NCI1} & $4{,}110$ & $2$ & $29.87$ & $32.30$ & $37$ \\
    \textbf{IMDB-BIN} & $1{,}000$ & $2$ & $19.77$ & $96.53$ & $-$ \\
    \textbf{IMDB-MUL} & $1{,}500$ & $3$ & $13.00$ & $65.94$ & $-$ \\
    \bottomrule
    \end{tabular}}
    \caption{Statistics of the data sets used in the empirical evaluation of the proposed models. We use three data sets with and two data sets without node features.}
    \label{tab:data_stats}
\end{table}

\subsection{Real-world experiments}

To examine whether the proposed GP models based on the energy profiles of attributed graphs are sufficiently expressive for classifying real-world graphs, we conduct a number of experiments on benchmark data sets and compare the model performance to popular baseline methods for graph classification.

The data sets in our empirical evaluation are shown in Table~\ref{tab:data_stats} along with an overview of their statistics.
The ENZYMES data set~\citep{borgwardt2005protein} contains protein graphs of enzymes that require classification into the Enzyme Commission top level enzyme classes.
The MUTAG data set~\citep{debnath1991structure} consists of molecular graphs and the task is to detect an effect on the Salmonella typhimurium bacterium.
Similarly, the NCI1 data set~\citep{wale2006comparison} holds molecular graphs that need to be classified based on whether they are active against certain types of cancer.
Finally, the IMDB-BINARY and IMDB-MULTI data sets~\citep{yanardag2015deep} are derived from the Internet Movie Database (\texttt{imdb.com}) and consist of graphs of actors and actresses who have co-starred together in a film. A classifier has to predict the genre of each graph belonging to a particular actor or actress.
The graphs in the first three data sets come with node attributes while the IMDB data sets are unattributed. We add one-hot encoded node degrees as features to ensure all graphs have node attributes.

We compare the proposed GP models to a number of neural network and graph kernel baselines.
The neural network baselines include DGCNN~\citep{zhang2018end}, GraphSAGE~\citep{hamilton2017inductive}, DiffPool~\citep{ying2018hierarchical}, and GIN~\citep{xu2019powerful}.
Among graph kernels we compare to the Shortest Path (SP) kernel~\citep{borgwardt2005shortest}, the Weisfeiler-Lehman (WL) kernel~\citep{shervashidze2011weisfeiler}, and the Multiscale Laplacian (ML) kernel~\citep{kondor2016multiscale}.

\begin{table*}
    \centering
    \scalebox{0.85}{
    \begin{tabular}{lccccc}
    \toprule
     & \textbf{ENZYMES} & \textbf{MUTAG} & \textbf{NCI1} & \textbf{IMDB-BINARY} & \textbf{IMDB-MULTI} \\
    \midrule
    \textbf{DGCNN} & \res{38.9}{5.7} & \res{84.0}{7.1} & \res{76.4}{1.7} & \res{69.2}{3.0} & \res{45.6}{3.4} \\
    \textbf{GraphSAGE} & \res{58.2}{6.0} & \res{83.6}{9.6} & \res{76.0}{1.8} & \res{68.8}{4.5} & \res{47.6}{3.5} \\
    \textbf{DiffPool} & \res{59.5}{5.6} & \res{79.8}{6.7} & \res{76.9}{1.9} & \res{68.4}{3.3} & \res{45.6}{3.4} \\
    \textbf{GIN} & \textcolor{highlightRed}{\res{\bd{59.6}}{4.5}} & \res{84.7}{6.7} & \textcolor{highlightOrange}{\res{\bd{80.0}}{1.4}} & \textcolor{highlightRed}{\res{\bd{71.2}}{3.9}} & \textcolor{highlightRed}{\res{\bd{48.5}}{3.3}} \\
    \midrule
    \textbf{SP} & timeout & \res{82.4}{5.5} & \res{72.5}{2.0} & \res{58.2}{4.7} & \res{39.2}{2.3} \\
    \textbf{WL} & \res{50.7}{7.3} & \textcolor{highlightRed}{\res{\bd{86.7}}{7.3}} & \textcolor{highlightBlue}{\res{\bd{85.2}}{2.2}} & \res{70.7}{6.8} & \textcolor{highlightBlue}{\res{\bd{51.3}}{4.4}} \\
    \textbf{ML} & \res{33.2}{5.8} & \textcolor{highlightOrange}{\res{\bd{87.2}}{7.5}} & \textcolor{highlightRed}{\res{\bd{79.7}}{1.8}} & \res{69.9}{4.8} & \res{47.7}{3.2} \\
    % \textbf{GIN} & \res{59.6}{4.5} & N/A & \res{80.0}{1.4} & \res{71.2}{3.9} & \res{48.5}{3.3} \\
    % \textbf{DiffPool} & \res{59.5}{5.6} & N/A & \res{76.9}{1.9} & \res{68.4}{3.3} & \res{45.6}{3.4} \\
    % \textbf{DGCNN} & \res{38.9}{5.7} & N/A & \res{76.4}{1.7} & \res{69.2}{3.0} & \res{45.6}{3.4} \\
    % \midrule
    % \textbf{3-WL} & N/A & \res{83.2}{0.0} & \res{00.0}{0.0} & \res{00.0}{0.0} & \res{00.0}{0.0} \\
    % \textbf{Graphlet} & \res{00.0}{0.0} & \res{00.0}{0.0} & \res{00.0}{0.0} & \res{00.0}{0.0} & \res{00.0}{0.0} \\
    % \textbf{SP} & \res{00.0}{0.0} & \res{00.0}{0.0} & \res{00.0}{0.0} & \res{00.0}{0.0} & \res{00.0}{0.0} \\
    \midrule
    \textbf{FT-GP} & \textcolor{highlightOrange}{\res{\bd{60.7}}{4.3}} & \res{85.7}{6.2} & \res{77.7}{1.6} & \textcolor{highlightOrange}{\res{\bd{72.7}}{3.9}} & \textcolor{highlightOrange}{\res{\bd{48.8}}{2.8}} \\
    \textbf{WT-GP} & \textcolor{highlightBlue}{\res{\bd{63.8}}{5.3}} & \textcolor{highlightBlue}{\res{\bd{87.3}}{4.8}} & \res{78.1}{2.1} & \textcolor{highlightBlue}{\res{\bd{74.6}}{4.1}} & \res{48.4}{2.9} \\
    \bottomrule
    \end{tabular}}
    \caption{Comparison of the proposed GP models with common neural network and graph kernel baselines in terms of classification accuracy. Colours indicate the \textcolor{highlightBlue}{best}, \textcolor{highlightOrange}{second-}, and \textcolor{highlightRed}{third-best} result.}
    \label{tab:results}
\end{table*}

The classification accuracy of each model on all data sets is shown in Table~\ref{tab:results} where the results for baseline methods are obtained from the empirical evaluation by~\citet{nikolentzos2021graph}.
Mirroring the results on the synthetic data set, we find that WT-GP performs at least as well as FT-GP on all data sets except on IMDB-MULTI where both perform roughly similarly.
Moreover, both models yield competitive performance compared to the baselines, with WT-GP achieving the highest accuracy on three of the five data sets and FT-GP being among the best two models also on three out of five data sets.
We highlight the surprising effectiveness of the Fourier features of FT-GP in comparison to the baselines, especially since no learning is involved in the feature construction.

While overall outperforming the neural network baselines, the trend across data sets is comparable between the GP models and the neural networks.
In fact, where the kernel methods outperform the GNNs, they also tend to be more comparable to the GP models.
This appears to provide evidence for the similarity between the methods proposed here and GNNs, as outlined in Section~\ref{sec:relationship}.

\subsection{Robustness analysis}\label{sec:robustness}

While the proposed GP models have fewer hyper-parameters than comparable neural network models, a small number of values needs to be selected prior to model fitting. For the FT-GP this is primarily the number of evaluation points $M$ and for WT-GP this is mainly the number of filters $K$.

We evaluate FT-GP for a range of different numbers of evaluation points $M$ on two data sets, one with and one without node features. 
The results are shown in Table~\ref{tab:robustness_eval_points}.
We find that FT-GP is overall robust to varying the number of evaluation points.
We note that larger $M$ linearly increase time and memory complexity for training and inference.
Notably, however, larger $M$ do not increase the number of learned parameters of FT-GP (which does not have learned parameters) but merely increase the ``resolution'' of the approximation of the cumulative energy function and we therefore do not expect FT-GP to begin to over-fit for larger $M$.

In a similar vein, the performance of WT-GP for varying numbers of filters $K$ is presented in Table~\ref{tab:robustness_filters}.
The results indicate that WT-GP is robust to different numbers of filters.
Similar to $M$, increasing $K$ linearly increases the time and memory complexity for training and inference.
Unlike $M$, however, increasing $K$ means a larger number of scale parameters need to be estimated using MLE and therefore we do expect the model to over-fit for very large values of $K$.
For the values that are computationally feasible for the given data sets, we have not yet observed over-fitting.

\begin{table}
\centering
\begin{tabular}{lcc}
\toprule
$M$ & \textbf{ENZYMES} & \textbf{IMDB-BINARY} \\
\midrule
$\bd{20}$ & \res{61.2}{5.1} & \res{73.1}{4.0} \\
$\bd{25}$ & \res{60.5}{4.5} & \res{73.5}{3.7} \\
$\bd{30}$ & \res{60.7}{4.3} & \res{72.7}{3.9} \\
$\bd{35}$ & \res{60.3}{4.9} & \res{73.6}{3.5} \\
$\bd{40}$ & \res{61.0}{4.9} & \res{73.0}{3.6} \\
$\bd{45}$ & \res{61.0}{4.9} & \res{73.0}{4.0} \\
$\bd{50}$ & \res{61.0}{4.7} & \res{72.9}{4.2} \\
\bottomrule
\end{tabular}
\caption{Performance of FT-GP on a data set with node features (ENZYMES) and a data set without node features (IMDB-BINARY) for varying number of evaluation points $M$.}
\label{tab:robustness_eval_points}
\end{table}

\begin{table}
\centering
\begin{tabular}{lcc}
\toprule
$K$ & \textbf{ENZYMES} & \textbf{IMDB-BINARY} \\
\midrule
$\bd{5}$ & \res{64.3}{5.0} & \res{74.6}{3.8} \\
$\bd{10}$ & \res{63.8}{5.3} & \res{74.6}{4.1} \\
$\bd{15}$ & \res{64.8}{5.4} & \res{74.3}{3.0} \\
$\bd{20}$ & \res{65.5}{5.5} & \res{74.2}{4.1} \\
\bottomrule
\end{tabular}
\caption{Performance of WT-GP on a data set with node features (ENZYMES) and a data set without node features (IMDB-BINARY) for varying number of filters $K$.}
\label{tab:robustness_filters}
\end{table}

% \subsection{Large-scale data sets}

\subsection{Quality of uncertainty estimates}\label{sec:uncertainty}

\begin{figure}
\centering
\textbf{FT-GP}\\
\vspace{0.15cm}
    \begin{subfigure}[b]{0.49\linewidth}
    \includegraphics[width=\linewidth]{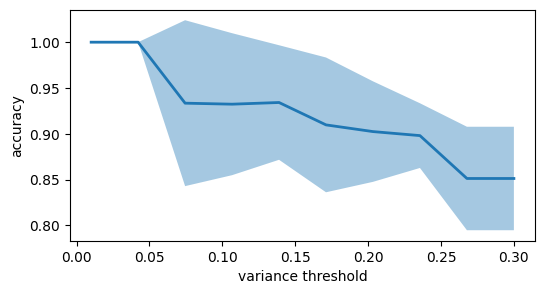}
    \caption{MUTAG}
    \end{subfigure}
    \begin{subfigure}[b]{0.49\linewidth}
    \includegraphics[width=\linewidth]{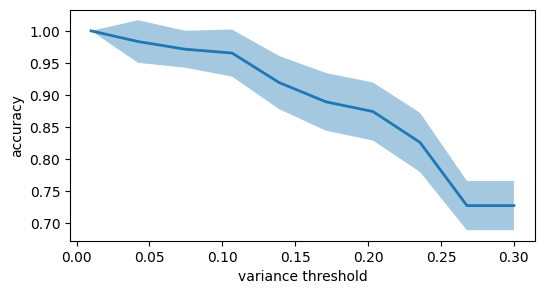}
    \caption{IMDB-BINARY}
    \end{subfigure}

\vspace{0.15cm}
\textbf{WT-GP}\\
\vspace{0.15cm}
    \begin{subfigure}[b]{0.49\linewidth}
    \includegraphics[width=\linewidth]{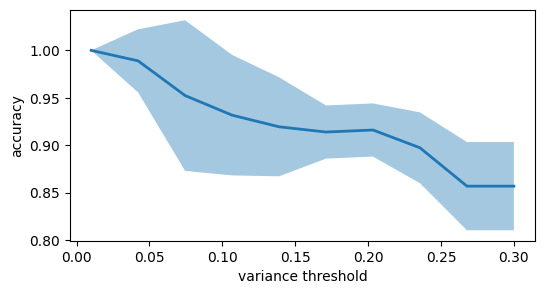}
    \caption{MUTAG}
    \end{subfigure}
    \begin{subfigure}[b]{0.49\linewidth}
    \includegraphics[width=\linewidth]{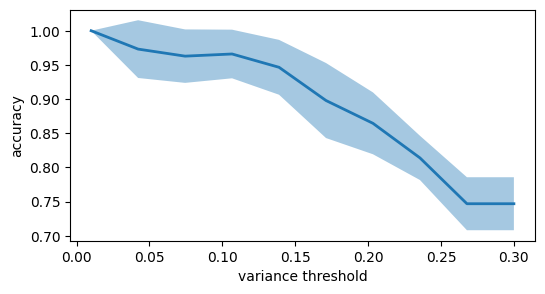}
    \caption{IMDB-BINARY}
    \end{subfigure}
\caption{Accuracy of predictions when rejecting predictions that have a variance above a certain threshold. As the threshold becomes lower, i.e. more and more strict (from right to left on x-axis), the prediction accuracy increases.
}\label{fig:uncertainty}
\end{figure}

One of the core advantages of a Bayesian treatment of graph classification is the availability of uncertainty estimates that can be a crucial part of the real-world application of a model.
Uncertainty estimates allow downstream users of the model to weigh the reliability of its predictions and thus more confidently make decisions based on those predictions.
The GP models proposed here make uncertainty predictions in the form of the variance of the variational posterior predictive distribution. 
We can assess the quality of those uncertainty estimates by simulating an experiment where downstream users are allowed to ``reject'' predictions if their variance is above a certain threshold.
We can then plot the accuracy of the predictions that are not rejected.
If the uncertainty estimates are well calibrated, we expect the accuracy of the remaining, low-variance (i.e. high certainty) predictions to increase. 
We plot the results for both our models and two of the data sets in Figure~\ref{fig:uncertainty}.
We find that, as expected, the accuracy increases for stricter variance thresholds. 
As the variance threshold increases, the accuracy rises from the results reported in Table~\ref{tab:results} for the complete set of predictions, to $1.0$ for only the most confident predictions.

\section{Conclusions}

We have proposed two GP models using spectral features to classify graphs.
The first approach constructs the energy profile of an attributed graph and compares across features.
We find that even though the model requires no learning to obtain these spectral features, it performs competitively to graph neural network and kernel baselines.
The second approach learns more complex wavelet filters and compares graphs based on the corresponding filtered node features.
It outperforms the first approach both on real-world and synthetic data sets.
% * Find both models produce useful uncertainty estimates
Our work indicates that spectral features constitute a powerful basis for graph classification both on their own and even more so when combined with the learning capabilities of GPs.

% \begin{contributions} % will be removed in pdf for initial submission 
% 					  % (without ‘accepted’ option in \documentclass)
%                       % so you can already fill it to test with the
%                       % ‘accepted’ class option
%     Briefly list author contributions. 
%     This is a nice way of making clear who did what and to give proper credit.
%     This section is optional.

%     H.~Q.~Bovik conceived the idea and wrote the paper.
%     Coauthor One created the code.
%     Coauthor Two created the figures.
% \end{contributions}

\begin{acknowledgements} % will be removed in pdf for initial submission,
						 % (without ‘accepted’ option in \documentclass)
                         % so you can already fill it to test with the
                         % ‘accepted’ class option
FLO acknowledges funding from the Huawei Hisilicon Studentship at the Department of Computer Science and Technology of the University of Cambridge. X.D. acknowledges support from the Oxford-Man Institute of Quantitative Finance and the EPSRC (EP/T023333/1).
\end{acknowledgements}

% References
\bibliography{opolka_465}

\begin{thebibliography}{45}
\providecommand{\natexlab}[1]{#1}
\providecommand{\url}[1]{\texttt{#1}}
\expandafter\ifx\csname urlstyle\endcsname\relax
  \providecommand{\doi}[1]{doi: #1}\else
  \providecommand{\doi}{doi: \begingroup \urlstyle{rm}\Url}\fi

\bibitem[Borgwardt et~al.(2020)Borgwardt, Ghisu, Llinares-L{\'o}pez, O’Bray,
  Rieck, et~al.]{borgwardt2020graph}
Karsten Borgwardt, Elisabetta Ghisu, Felipe Llinares-L{\'o}pez, Leslie
  O’Bray, Bastian Rieck, et~al.
\newblock Graph kernels: State-of-the-art and future challenges.
\newblock \emph{Foundations and Trends{\textregistered} in Machine Learning},
  13\penalty0 (5-6), 2020.

\bibitem[Borgwardt and Kriegel(2005)]{borgwardt2005shortest}
Karsten~M. Borgwardt and Hans-Peter Kriegel.
\newblock Shortest-path kernels on graphs.
\newblock In \emph{Fifth IEEE International Conference on Data Mining}, 2005.

\bibitem[Borgwardt et~al.(2005)Borgwardt, Ong, Schönauer, Vishwanathan, Smola,
  and Kriegel]{borgwardt2005protein}
Karsten~M. Borgwardt, Cheng~Soon Ong, Stefan Schönauer, S.~V.~N. Vishwanathan,
  Alex~J. Smola, and Hans-Peter Kriegel.
\newblock {Protein function prediction via graph kernels}.
\newblock \emph{Bioinformatics}, 21, 2005.

\bibitem[Bruna et~al.(2014)Bruna, Zaremba, Szlam, and LeCun]{bruna2013spectral}
Joan Bruna, Wojciech Zaremba, Arthur Szlam, and Yann LeCun.
\newblock Spectral networks and locally connected networks on graphs.
\newblock In \emph{2nd International Conference on Learning Representations},
  2014.

\bibitem[Byrd et~al.(1995)Byrd, Lu, Nocedal, and Zhu]{byrd1995limited}
Richard~H. Byrd, Peihuang Lu, Jorge Nocedal, and Ciyou Zhu.
\newblock A limited memory algorithm for bound constrained optimization.
\newblock \emph{SIAM Journal on Scientific Computing}, 16\penalty0 (5), 1995.

\bibitem[Chung(1997)]{chung1997spectral}
Fan R.~K. Chung.
\newblock \emph{Spectral Graph Theory}.
\newblock American Mathematical Society, 1997.

\bibitem[Dai et~al.(2016)Dai, Dai, and Song]{dai2016discriminative}
Hanjun Dai, Bo~Dai, and Le~Song.
\newblock Discriminative embeddings of latent variable models for structured
  data.
\newblock In \emph{Proceedings of the 33rd International Conference on Machine
  Learning}, 2016.

\bibitem[Debnath et~al.(1991)Debnath, Lopez~de Compadre, Debnath, Shusterman,
  and Hansch]{debnath1991structure}
Asim~Kumar Debnath, Rosa~L. Lopez~de Compadre, Gargi Debnath, Alan~J.
  Shusterman, and Corwin Hansch.
\newblock Structure-activity relationship of mutagenic aromatic and
  heteroaromatic nitro compounds. correlation with molecular orbital energies
  and hydrophobicity.
\newblock \emph{Journal of Medicinal Chemistry}, 34\penalty0 (2), 1991.

\bibitem[Defferrard et~al.(2016)Defferrard, Bresson, and
  Vandergheynst]{defferrard2016chebnet}
Micha\"{e}l Defferrard, Xavier Bresson, and Pierre Vandergheynst.
\newblock Convolutional neural networks on graphs with fast localized spectral
  filtering.
\newblock In \emph{Advances in Neural Information Processing Systems}, 2016.

\bibitem[Dong et~al.(2013)Dong, Ortega, Frossard, and
  Vandergheynst]{dong2013inference}
Xiaowen Dong, Antonio Ortega, Pascal Frossard, and Pierre Vandergheynst.
\newblock Inference of mobility patterns via spectral graph wavelets.
\newblock In \emph{2013 IEEE International Conference on Acoustics, Speech and
  Signal Processing}, 2013.

\bibitem[Dong et~al.(2020)Dong, Thanou, Toni, Bronstein, and
  Frossard]{dong2020graph}
Xiaowen Dong, Dorina Thanou, Laura Toni, Michael~M. Bronstein, and Pascal
  Frossard.
\newblock Graph signal processing for machine learning: A review and new
  perspectives.
\newblock \emph{IEEE Signal Processing Magazine}, 37\penalty0 (6), 2020.

\bibitem[Duvenaud et~al.(2015)Duvenaud, Maclaurin, Iparraguirre, Bombarell,
  Hirzel, Aspuru-Guzik, and Adams]{duvenaud2015convolutional}
David~K Duvenaud, Dougal Maclaurin, Jorge Iparraguirre, Rafael Bombarell,
  Timothy Hirzel, Al{\'a}n Aspuru-Guzik, and Ryan~P Adams.
\newblock Convolutional networks on graphs for learning molecular fingerprints.
\newblock In \emph{Advances in neural information processing systems}, 2015.

\bibitem[Erd\"{o}s and R\'{e}nyi(1959)]{erdos1959random}
Paul Erd\"{o}s and Alfr\'{e}d R\'{e}nyi.
\newblock On random graphs {I}.
\newblock \emph{Publicationes Mathematicae Debrecen}, 1959.

\bibitem[Errica et~al.(2020)Errica, Podda, Bacciu, and Micheli]{errica2020fair}
Federico Errica, Marco Podda, Davide Bacciu, and Alessio Micheli.
\newblock A fair comparison of graph neural networks for graph classification.
\newblock In \emph{8th International Conference on Learning Representations},
  2020.

\bibitem[Gilmer et~al.(2017)Gilmer, Schoenholz, Riley, Vinyals, and
  Dahl]{gilmer2017neural}
Justin Gilmer, Samuel~S. Schoenholz, Patrick~F. Riley, Oriol Vinyals, and
  George~E. Dahl.
\newblock Neural message passing for quantum chemistry.
\newblock In \emph{Proceedings of the 34th International Conference on Machine
  Learning}, 2017.

\bibitem[Hamilton et~al.(2017)Hamilton, Ying, and
  Leskovec]{hamilton2017inductive}
Will Hamilton, Zhitao Ying, and Jure Leskovec.
\newblock Inductive representation learning on large graphs.
\newblock In \emph{Advances in neural information processing systems}, 2017.

\bibitem[Hammond et~al.(2011)Hammond, Vandergheynst, and
  Gribonval]{hammond2011wavelets}
David~K Hammond, Pierre Vandergheynst, and R{\'e}mi Gribonval.
\newblock Wavelets on graphs via spectral graph theory.
\newblock \emph{Applied and Computational Harmonic Analysis}, 30\penalty0 (2),
  2011.

\bibitem[Hensman et~al.(2013)Hensman, Fusi, and Lawrence]{hensman2013gaussian}
James Hensman, Nicol\`{o} Fusi, and Neil~D. Lawrence.
\newblock Gaussian processes for big data.
\newblock In \emph{Proceedings of the 29th Conference on Uncertainty in
  Artificial Intelligence}, 2013.

\bibitem[Hensman et~al.(2015)Hensman, de~G.~Matthews, and
  Ghahramani]{hensman2015scalable}
James Hensman, Alexander~G. de~G.~Matthews, and Zoubin Ghahramani.
\newblock Scalable variational gaussian process classification.
\newblock In \emph{Proceedings of the 18th International Conference on
  Artificial Intelligence and Statistics}, 2015.

\bibitem[Holland et~al.(1983)Holland, Laskey, and
  Leinhardt]{holland1983stochastic}
Paul~W. Holland, Kathryn~Blackmond Laskey, and Samuel Leinhardt.
\newblock Stochastic blockmodels: First steps.
\newblock \emph{Social Networks}, 5\penalty0 (2), 1983.

\bibitem[Huang and Villar(2021)]{huang2021short}
Ningyuan~Teresa Huang and Soledad Villar.
\newblock A short tutorial on the weisfeiler-lehman test and its variants.
\newblock In \emph{IEEE International Conference on Acoustics, Speech and
  Signal Processing}, 2021.

\bibitem[Kipf and Welling(2017)]{kipf2017semi}
Thomas~N. Kipf and Max Welling.
\newblock Semi-supervised classification with graph convolutional networks.
\newblock In \emph{5th International Conference on Learning Representations},
  2017.

\bibitem[Kondor and Pan(2016)]{kondor2016multiscale}
Risi Kondor and Horace Pan.
\newblock The multiscale laplacian graph kernel.
\newblock In \emph{Advances in neural information processing systems}, 2016.

\bibitem[Kriege et~al.(2020)Kriege, Johansson, and Morris]{kriege2020survey}
Nils~M Kriege, Fredrik~D Johansson, and Christopher Morris.
\newblock A survey on graph kernels.
\newblock \emph{Applied Network Science}, 5\penalty0 (1), 2020.

\bibitem[Le~Magoarou et~al.(2018)Le~Magoarou, Gribonval, and
  Tremblay]{lemagoarou2018approximate}
Luc Le~Magoarou, Rémi Gribonval, and Nicolas Tremblay.
\newblock Approximate fast graph fourier transforms via multilayer sparse
  approximations.
\newblock \emph{IEEE Transactions on Signal and Information Processing over
  Networks}, 4\penalty0 (2), 2018.

\bibitem[Li et~al.(2012)Li, Zhu, Chi, and Zhang]{li2012nested}
Bin Li, Xingquan Zhu, Lianhua Chi, and Chengqi Zhang.
\newblock Nested subtree hash kernels for large-scale graph classification over
  streams.
\newblock In \emph{2012 IEEE 12th International Conference on Data Mining},
  2012.

\bibitem[{Liu} and {Nocedal}(1989)]{liu1989limited}
Dong~C. {Liu} and Jorge {Nocedal}.
\newblock On the limited memory {B}{F}{G}{S} method for large scale
  optimization.
\newblock \emph{Mathematical Programming}, 45\penalty0 (3, (Ser. B)), 1989.

\bibitem[Murphy et~al.(2019)Murphy, Srinivasan, Rao, and
  Ribeiro]{murphy2019relational}
Ryan Murphy, Balasubramaniam Srinivasan, Vinayak Rao, and Bruno Ribeiro.
\newblock Relational pooling for graph representations.
\newblock In \emph{Proceedings of the 36th International Conference on Machine
  Learning}, 2019.

\bibitem[Nikolentzos et~al.(2021)Nikolentzos, Siglidis, and
  Vazirgiannis]{nikolentzos2021graph}
Giannis Nikolentzos, Giannis Siglidis, and Michalis Vazirgiannis.
\newblock Graph kernels: A survey.
\newblock \emph{Journal of Artificial Intelligence Research}, 72, 2021.

\bibitem[Opolka et~al.(2022)Opolka, Zhi, Li\`o, and Dong]{opolka2022adaptive}
Felix Opolka, Yin-Cong Zhi, Pietro Li\`o, and Xiaowen Dong.
\newblock Adaptive gaussian processes on graphs via spectral graph wavelets.
\newblock In \emph{Proceedings of The 25th International Conference on
  Artificial Intelligence and Statistics}, 2022.

\bibitem[Ortega(2022)]{ortega2022introduction}
Antonio Ortega.
\newblock \emph{Introduction to Graph Signal Processing}.
\newblock Cambridge University Press, 2022.

\bibitem[Ortega et~al.(2018)Ortega, Frossard, Kova{\v{c}}evi{\'c}, Moura, and
  Vandergheynst]{ortega2018graph}
Antonio Ortega, Pascal Frossard, Jelena Kova{\v{c}}evi{\'c}, Jos{\'e}~MF Moura,
  and Pierre Vandergheynst.
\newblock Graph signal processing: Overview, challenges, and applications.
\newblock \emph{Proceedings of the IEEE}, 106\penalty0 (5), 2018.

\bibitem[Pineau(2019)]{pineau2019using}
Edouard Pineau.
\newblock Using laplacian spectrum as graph feature representation.
\newblock \emph{arXiv preprint arXiv:1912.00735}, 2019.

\bibitem[Porter and Canagarajah(1996)]{porter1996robust}
Robert Porter and Nishan Canagarajah.
\newblock A robust automatic clustering scheme for image segmentation using
  wavelets.
\newblock \emph{IEEE transactions on image processing}, 5\penalty0 (4), 1996.

\bibitem[Rasmussen and Williams(2005)]{rasmussen2005gp}
Carl~Edward Rasmussen and Christopher K.~I. Williams.
\newblock \emph{Gaussian Processes for Machine Learning (Adaptive Computation
  and Machine Learning)}.
\newblock The MIT Press, 2005.

\bibitem[Shervashidze et~al.(2011)Shervashidze, Schweitzer, van Leeuwen,
  Mehlhorn, and Borgwardt]{shervashidze2011weisfeiler}
Nino Shervashidze, Pascal Schweitzer, Erik~Jan van Leeuwen, Kurt Mehlhorn, and
  Karsten~M. Borgwardt.
\newblock Weisfeiler-lehman graph kernels.
\newblock \emph{Journal of Machine Learning Research}, 12, 2011.

\bibitem[Shuman et~al.(2013)Shuman, Narang, Frossard, Ortega, and
  Vandergheynst]{shuman2012emerging}
David~I Shuman, Sunil~K Narang, Pascal Frossard, Antonio Ortega, and Pierre
  Vandergheynst.
\newblock The emerging field of signal processing on graphs: Extending
  high-dimensional data analysis to networks and other irregular domains.
\newblock \emph{IEEE signal processing magazine}, 30\penalty0 (3), 2013.

\bibitem[Titsias(2009)]{titsias2009variational}
Michalis Titsias.
\newblock Variational learning of inducing variables in sparse gaussian
  processes.
\newblock In \emph{Proceedings of the 12th International Conference on
  Artificial Intelligence and Statistics}, 2009.

\bibitem[Togninalli et~al.(2019)Togninalli, Ghisu, Llinares-L{\'o}pez, Rieck,
  and Borgwardt]{togninalli2019wasserstein}
Matteo Togninalli, Elisabetta Ghisu, Felipe Llinares-L{\'o}pez, Bastian Rieck,
  and Karsten Borgwardt.
\newblock Wasserstein weisfeiler-lehman graph kernels.
\newblock \emph{Advances in Neural Information Processing Systems}, 32, 2019.

\bibitem[Wale and Karypis(2006)]{wale2006comparison}
Nikil Wale and George Karypis.
\newblock Comparison of descriptor spaces for chemical compound retrieval and
  classification.
\newblock In \emph{Sixth International Conference on Data Mining}, 2006.

\bibitem[Wu et~al.(2019)Wu, Souza, Zhang, Fifty, Yu, and
  Weinberger]{wu2019simplifying}
Felix Wu, Amauri Souza, Tianyi Zhang, Christopher Fifty, Tao Yu, and Kilian
  Weinberger.
\newblock Simplifying graph convolutional networks.
\newblock In \emph{Proceedings of the 36th International Conference on Machine
  Learning}, 2019.

\bibitem[Xu et~al.(2019)Xu, Hu, Leskovec, and Jegelka]{xu2019powerful}
Keyulu Xu, Weihua Hu, Jure Leskovec, and Stefanie Jegelka.
\newblock How powerful are graph neural networks?
\newblock In \emph{7th International Conference on Learning Representations},
  2019.

\bibitem[Yanardag and Vishwanathan(2015)]{yanardag2015deep}
Pinar Yanardag and S.V.N. Vishwanathan.
\newblock Deep graph kernels.
\newblock In \emph{Proceedings of the 21th ACM SIGKDD International Conference
  on Knowledge Discovery and Data Mining}, 2015.

\bibitem[Ying et~al.(2018)Ying, You, Morris, Ren, Hamilton, and
  Leskovec]{ying2018hierarchical}
Rex Ying, Jiaxuan You, Christopher Morris, Xiang Ren, William~L. Hamilton, and
  Jure Leskovec.
\newblock Hierarchical graph representation learning with differentiable
  pooling.
\newblock In \emph{Advances in Neural Information Processing Systems}, 2018.

\bibitem[Zhang et~al.(2018)Zhang, Cui, Neumann, and Chen]{zhang2018end}
Muhan Zhang, Zhicheng Cui, Marion Neumann, and Yixin Chen.
\newblock An end-to-end deep learning architecture for graph classification.
\newblock \emph{Proceedings of the AAAI Conference on Artificial Intelligence},
  32\penalty0 (1), 2018.

\end{thebibliography}
\end{document}